\documentclass[runningheads]{llncs}
\usepackage{graphicx}
\usepackage{algorithm}
\usepackage{algpseudocode}
\usepackage{amssymb}
\usepackage{amsmath}
\usepackage{xcolor}
\usepackage{dcolumn}
\usepackage{tabu}
\usepackage{xspace}
\usepackage[caption=false]{subfig}
\usepackage{multirow}
\usepackage{hyperref}

\newcommand{\GTrans}{\textsc{GTrans}\xspace}
\newcommand{\etal}[0]{\textit{et al.}\xspace}

\def\preview{1}
\def\submit{0}

\if\preview0
\usepackage[colorinlistoftodos]{todonotes}
\newcounter{todocounter}

\else
\usepackage[colorinlistoftodos,disable]{todonotes}
\newcounter{todocounter}

\fi

\titlerunning{Spatiotemporal Transformer with Graph Embeddings}
\title{\GTrans: Spatiotemporal Autoregressive Transformer with Graph Embeddings for Nowcasting Extreme Events}

\if\submit1
    \author{Anonymous}
    \authorrunning{Anonymous et al.}
    \institute{Anonymous institute}
    \authorrunning{Anonymous et al.}
\else
    \author{
    Bo Feng\inst{1} \and
    Geoffrey Fox\inst{2}
    }
    \authorrunning{Feng and Fox}
    \institute{Luddy School of Informatics, Computing, and Engineering\\
    Indiana University Bloomington, Bloomington IN, USA\\
    \email{fengbo@iu.edu}\\
    \and
    Biocomplexity Institute and Initiative and Computer Science Department,
    University of Virginia, Charlottesville, VA, USA\\
    \email{gcfexchange@gmail.com}}
\fi

\begin{document}

\maketitle              
\begin{abstract}
Spatiotemporal time series nowcasting should preserve temporal and spatial dynamics in the sense that generated new sequences from models respect the covariance relationship from history. Conventional feature extractors are built with deep convolutional neural networks (CNN). 
However, CNN models have limits to image-like applications where data can be formed with high-dimensional arrays. In contrast, applications in social networks, road traffic, physics, and chemical property prediction where data features can be organized with nodes and edges of graphs. 
Transformer architecture is an emerging method for predictive models, bringing high accuracy and efficiency due to attention mechanism design. This paper proposes a spatiotemporal model, namely \GTrans, that transforms data features into graph embeddings and predicts temporal dynamics with a transformer model. According to our experiments, we demonstrate that \GTrans can model spatial and temporal dynamics and nowcasts extreme events for datasets. Furthermore, in all the experiments, \GTrans can achieve the highest F$_1$ and F$_2$ scores in binary-class prediction tests than the baseline models.

\end{abstract}
\section{Introduction}
\label{sec:intro}

In time series machine learning, extreme events forecasting is a crucial subset interested in forecasting variables of interest that have some rare probability. Spatial and temporal attributes have played an essential role in addressing scientific issues mathematically and statistically with large volumes of data in real problems. 

Conventional feature extractors are built with deep convolutional neural networks (CNN). In contrast to applying CNN to image-like applications, graphs in machine learning are widely adopted in various applications in social networks, road traffic, physics, and chemical property prediction, where data features can be modeled with nodes and edges of graphs. Geometric deep learning aims to model non-Euclidean domains, which is suitable for many scientific applications~\cite{bronstein2017geometric}. Moreover, deep graph convolution neural networks can learn graph node embeddings~\cite{kipf2016semi}.

Recurrent neural networks are the de-facto in building predictive models. However, they are inefficient in training. Transformer architecture~\cite{vaswani2017attention,dosovitskiy2020image,zerveas2021transformer} is an emerging method for predictive models. Due to its originality of designing for natural language processing, there is a gap bridging the machine learning problem settings to the capability of Transformers.

Nowcasting for extreme events is a particular subset of problems similar to anonymous detection. However, unlike anonymous detection problems such as detecting a fraud credit card transaction, it is instantly known if an event is extreme by given features. For example, it is easy to tell an earthquake with a high magnitude, a severe accident if persons get killed in a traffic collision or severe weather condition. So, self-supervision or un-supervision is required to model this set of problems.

Inspired by the multi-head attention mechanism and encoder-decoder architecture, in this paper, we introduce an advanced transformer-based neural network autoencoder that encodes static spatial information, temporal information, decodes the latent space representation to future features in unsupervised learning tasks, and predicts quantile measures of probabilities for extreme events-of-interest. Self-supervised learning follows the forecasting nature that historical information is the most predictive helpful information for future values.
The contribution of this paper is:
\begin{itemize}
    \item We model the spatial features with graphs and encode graphs with graph convolutional neural networks.
    \item We integrate the graph embeddings with the Transformer encoder and decoder to build a predictive model for nowcasting extreme events.
\end{itemize}

The remainder of this paper is organized as follows. Section~\ref{sec:rel} summarizes related work in learning graph representation and time-series forecasting models. In Section~\ref{sec:modeling}, we formalize the nowcasting problem.
Section~\ref{sec:arch} illustrates the proposed model architecture. We test and analyze the model in Section~\ref{sec:exp}. Finally, we conclude the paper in Section~\ref{sec:con}.

\section{Related Work}
\label{sec:rel}

\subsection{Graph Representation Learning}

There are various methods have been proposed to estimate the graph representations.
Kipf and Welling~\cite{kipf2016semi} proposed a simplified graph convolutional network to classify graph nodes from local neighbors. 
Yan \etal proposed a Spatial Temporal Graph Convolutional Networks (STGCN)
to model dynamic skeletons~\cite{yan2018spatial}. In this paper, a graph is formed as nodes
defined in body joints and edges based on bones or natural connections in human
bodies. A spatial temporal graph is a skeleton sequence of body actions. This
is a classification problem in which the output of the model is categorized as a
type of movement such as running or jumping.
Geng \etal~\cite{geng2019spatiotemporal} uses multi-graphs to model regional ride-hailing demand. In their work, multiple-graphs are used to model spatial information and  gated recurrent neural network for temporal correlation.

\subsection{Time Series Predictive Models}

DeepAR is a symmetric encoder decoder neural network that predicts the probabilistic~\cite{salinas2020deepar}.
TimeGAN~\cite{yoon2019time} is a generative adversarial network (GAN) for time series data, which consist of static features and temporal features. In TimeGAN, there is an embedding function, recovery function, sequence generator, and sequence discriminator.
TransE~\cite{bordes2013translating} is one of the earliest successful project that predict data relationship using embeddings. Successors adopt to use advanced models for learning embeddings and predictive model for temporal dynamics. 
Zhu \etal~\cite{zhu2020adversarial} develop another GAN using one LSTM network as the generator and another LSTM network as the discriminator. Compared to GAN-based models, our model adopts GCN for embedding generation and Transformer for capturing temporal dynamics. Wu \etal demonstrated a successful application of using Transformer to prediction influenza prevalence~\cite{wu2020deep}.
Zerveas \etal~\cite{zerveas2021transformer} proposed a transformer-based framework for learing time series representation. As opposed to the original transformer from Vaswani \etal~\cite{vaswani2017attention}, they drop the decoder module but adopt a unified multi-variate feature matrix.

\section{Problem Modeling Overview}
\label{sec:modeling}

\subsection{Problem Formulation}
Consider a spatiotemporal time series data setting where each sample is an event $\mathbf{x}^i_t = \{s_i, x_t\}$ which consists of two elements: spatial features $s_i$ at location $i$ and temporal features $x_t$ at time $t$.

Define $s_i$  as $\{s_i: \mathbb{R}^m, 1 \leq i \leq n \}$, $x_t$ as a vector from space $\{ \mathbb{R}^k \}$ at time $t$. Let $T$ be a constant length of look-back time steps. A history window of sequence at given time $t$ is $\{ E_{t-T}, E_{t-T+1}, \dots, E_{t} \}$, and its known future sequence is $\{ \mathbf{x}_{t-T+1}, \mathbf{x}_{t-T+2}, \dots, \mathbf{x}_{t+1} \}$ in the training phase.
For the convenience, we denote $\mathbf{x}_t$ as the input signal consisting of all spatial and temporal features at time $t$.
\begin{equation}
    \operatorname*{min}_{ \hat{p} }D \Big(p(\mathbf{x}_t | \mathbf{x}_{t-T:t-1}) || \hat{p}(\mathbf{x}_t | \mathbf{x}_{t-T:t-1}) \Big),
\end{equation}
where $\hat{p}(E_t | E_{t-T:t-1})$ is an estimate joint density function that approximates $p(E_t | E_{t-T:t-1})$ at any given time $t$.

\subsection{Graph Embedding with Laplacian Smoothing and Sharpening}
\label{sec:graph-embedding}
Given a graph is defined as $G = (V, E, A)$, where $V$ and $E$ denotes vertices and edges respectively, $A \in \mathbb{R}^{n \times n}$ is the adjacency matrix of the graph with $n$ nodes, which encodes the pairwise distance between nodes. The symmetric graph laplacian is then calculated as
\begin{equation}
    L = I_n - D^{-\frac{1}{2}}AD^{-\frac{1}{2}},
\end{equation}
where $L, I_n \in \mathbb{R}^{n \times n}$, $I_n$ is an identity matrix, and $D$ is a diagonal matrix, in which $D_{ii}=\sum_{j}A_{ij}$.

Laplacian smoothing creates a weighted average representations from local inputs and neighbors~\cite{taubin1995signal}. For each index $i$ of node features, the Laplacian smoothing is defined as follows:
\begin{equation}\label{eq:smoothing}
    x_i^{(m+1)} = (1-\gamma)x_i^{(m)} + \gamma\sum_{j}\frac{\tilde{A}_{ij}}{\tilde{D}_{ii}}x_j^{(m)},
\end{equation}
where $x_i^{(m+1)}$ is the reconstructed features from $x_i^{(m)}$.

Laplacian sharpening is a reversed process that destructs the reconstructed node features from centroids of neighbors. This process is expressed as a similar form as Eq~\ref{eq:smoothing} as follows:
\begin{equation}\label{eq:sharpening}
    x_i^{(m+1)} = (1+\gamma)x_i^{(m)} - \gamma\sum_{j}\frac{\tilde{A}_{ij}}{\tilde{D}_{ii}}x_j^{(m)},
\end{equation}
where $x_i^{(m+1)}$ is the deconstructed features from $x_i^{(m)}$.

\subsection{Long Term Series Nowcasting for Extreme Events}

Inspired by \cite{wang2018non}, here we define the long term series nowcasting in formulation as:
\begin{equation}
    \mathbf{y}_j = \mathcal{D}( \frac{\sum_{\forall i} f(\mathbf{x}_i, \mathbf{x}_i) \mathcal{G}(\mathbf{x}_j)}{\mathcal{C}(\mathbf{x})} )
\end{equation}
, where $\mathbf{x}_i$ is the time series in the input window and $\mathbf{x}_j$ is the output time series in the next window. $\mathcal{G}$ is the graph encoding function and $\mathcal{D}$ is the graph decoding function. A function $f$ computes the time-wise attention from $i$ to $j$. The temporal correlation is normalized by a factor $\mathcal{C}$. This formulae states that $\mathbf{x}$, $\mathbf{y}$ share the same size of features but the order of relationship can be non-linear.

\section{\GTrans: Spatiotemporal Transformer with Graph Embeddings}
\label{sec:arch}

We describe the overall model architecture and components in this section. As opposed to related work, our model is able to learn representation for various of input data structures. Graph embeddings keep the spatial information between nodes and the transformer layers are able to carry predictive tasks.

\begin{figure}[htbp]
    \centering
    \includegraphics[width=\textwidth]{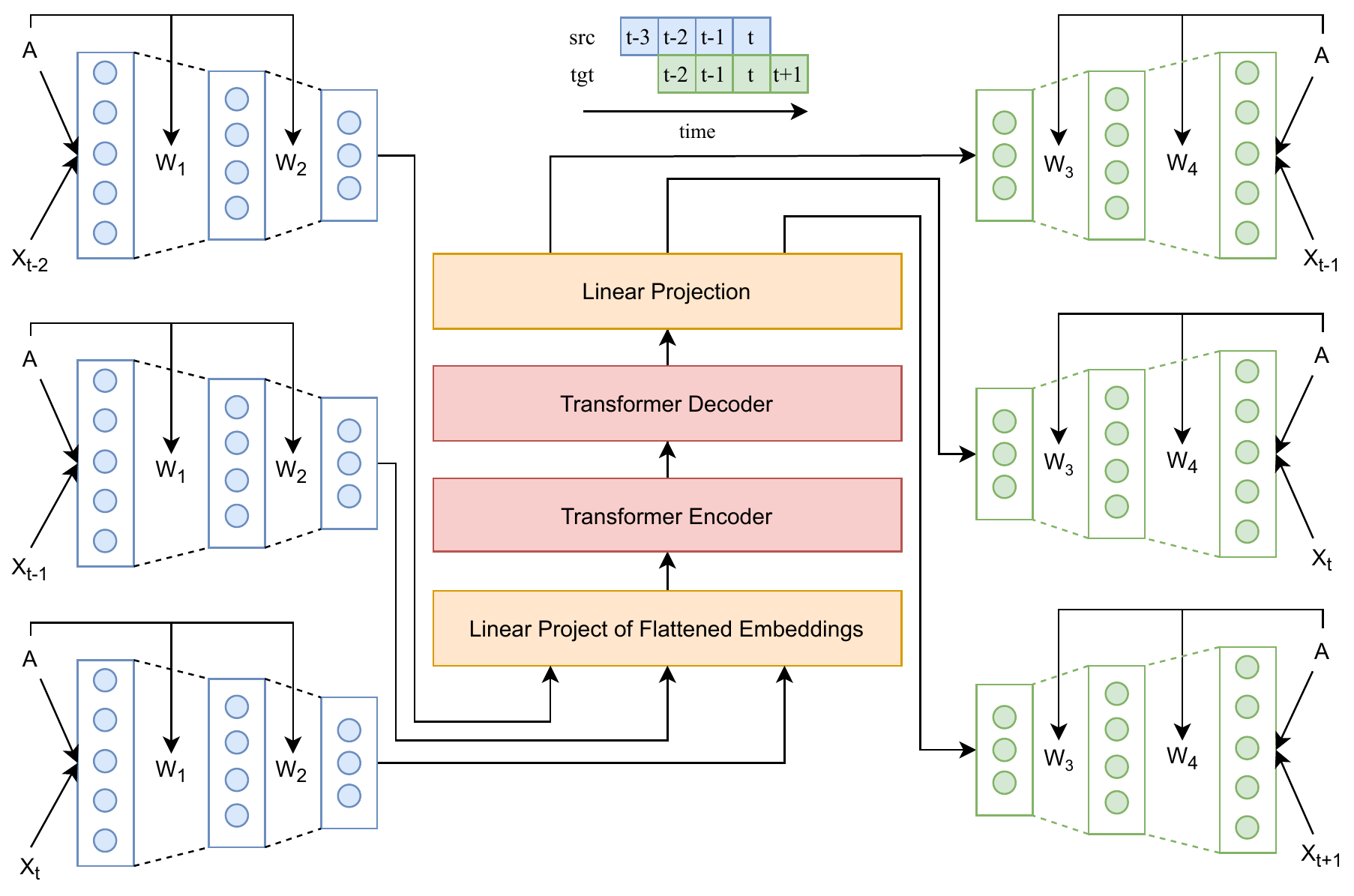}
    \caption{\GTrans architecture: Time distributed graph convolutional embeddings with Transformer encoder and decoder. This figure gives an example of using two convolutional layers in the graph encoder.}
    \label{fig:model-arch}
\end{figure}

\subsection{Spectral convolution on graphs}

A spectral convolution on a graph input is a multiplication of element-wise signal $x \in \mathbb{R}^n$ with its adjacency matrix. The spectral convolution is approximated as follows:
\begin{equation}
    g_{\theta} * x = \theta ( I_n + D^{-\frac{1}{2}} A D^{-\frac{1}{2}})x,
\end{equation}
where $A$ is the adjacency matrix and $D \in \mathbb{R}^{n\times n} $ is a diagonal matrix with $D_{ii}=\sum_j A_{ij}$.

For symmetric graph encoding and decoding, which we expect the input $X$ is close to the target $\hat{X}$, so the cost of spatial reconstruction is given by
\begin{equation}
    \frac{1}{T} \sum_{i=1}^{T} \min_{\hat{X}_i} \frac{1}{2} || X_i - \hat{X}_i ||^2_{F}
\end{equation}
where $T$ is the time window size and $F$ denotes any distance function.

\subsection{Symmetric Graph Convolutional Encoding and Decoding}

Graph convolutional modules are used to encode and decode spatial embeddings. In our proposed architecture (as shown in Figure~\ref{fig:model-arch}), the graph encoder and decoder are symmetric, which means the graph encoder and decoder contain the same number of layers and the layers' shapes are mirrored from encoding to decoding. This design allows the Laplacian smoothing in the encoder and Laplacian sharpening in the decoder while keeping the graph structures of node features~\cite{park2019symmetric}. As the embedding generation is stated in Section~\ref{sec:graph-embedding} (Eq.~\ref{eq:smoothing} and~\ref{eq:sharpening}), the graph encoder passes the embedding to the transformer ordered in time-wise. Then the transformer passes the same time-wise order to the graph decoder. The graph decoding is a reversed process of the encoding. In \GTrans, we adopt a simplified version of Graph encoding and decoding as used in~\cite{kipf2016semi,park2019symmetric}.

\subsection{Spatiotemporal Transformer AutoEncoder}

We depict an end-to-end spatiotemporal dependency modeling by connecting the Graph components and Transformer components.
Figure~\ref{fig:model-arch} illustrates the overall model architecture, which consists three main components 1) graph encoder, 2) transformer encoder, decoder, and 3) graph decoder.
The Graph encoder receives an input $ X \in \mathbb{R}^{T \times N \times C} $ and an adjacency matrix $A \in \mathbb{R}^{N\times N}$ that defines the spatial relationship between nodes. In this input, $T$ represents the time window size, $N$ is the number of nodes and $C$ is the feature size for each node. The generated embedding is aligned to the time axis with a shape as $T\times N \times D$, in which $D$ is the embedding dimension size. The transformer encoder takes the graph embeddings $E \in \mathbb{R}^{T\times N \times D} $ and produces a memory context with the shape as $ T \times N \times D $ as the same as input. By passing both the memory context and the input for the graph encoder, Transformer decoder predicts a future embedding $\hat{E} \in \mathbb{R}^{T\times N \times D} $. Then, this embedding is fed into the graph decoder, which decodes as the predicted output $ \hat{X} \in \mathbb{R}^{T \times N \times C} $ with the same adjacency matrix $A \in \mathbb{R}^{N\times N}$. We describe the details of Transformer components as follows:

\subsubsection{Positional Embedding}
The input for Transformer encoder is a flattened output of Graph encoder. So we project the input to $D$ dimensions with a trainable linear operation, which is similar to the positional mapping in ViT~\cite{dosovitskiy2020image} and~\cite{bertasius2021space} except that there is no class feature. So the positional embedding is calculated as:
\begin{equation}
    z_t = Ex_t + e^{pos}_t
\end{equation}, where $E$ is a projection weight matrix and $e^{pos}_t$ is a learnable linear positional embedding.

\subsubsection{Multi-Head Self-Attention}
We use the Multi-Head self-attention (MSA) as opposed to the standard self-attention mechanism (SA) as~\cite{vaswani2017attention} by splitting the long sequence size for each input embedding sequence $\mathbf{z} \in \mathbb{R}^{T\times (N\cdot D)}$ of every $T$ time steps. The SA and MSA are calculated as follows:
\begin{align}\label{eq:msa}
     [\mathbf{q},\mathbf{k},\mathbf{v}] &= \mathbf{z}\mathbf{H}_{qkv} \\
     Attn &= softmax(\frac{\mathbf{q}K^T}{\sqrt{D_h}}) \\
     SA(\mathbf{z}) &= Attn\mathbf{v} \\
     MSA(\mathbf{z}) &= [SAttn_1(z);SAttn_2(z);\dots;SAttn_k(z)] \mathbf{H}_{msa}
\end{align}
In these equations Eq.~\ref{eq:msa}, the attention weight $Attn \in \mathbb{R}^{T\times T}$ is a pair-wise matrix for every two input sequences.

\subsubsection{Temporal Projection}
Temporal projection is simply a process that aligns the decoded output to the following graph decoder. Given the transformer input and target $E \in \mathbb{R}^{T\times N \times D} $, the nowcasted embedding is then projected to $\hat{E} \in \mathbb{R}^{T\times N\times D}$.

Overall, the spatial information is retained in the graph embeddings and transformer components are responsible for capturing the temporal dynamics.

\subsection{Self-supervised Training}

The model is self-supervised by its input and recovered output. The modeling loss is defined as:
\begin{equation}\label{eq:loss}
    \frac{\lambda}{T} \sum_{i=1}^{T} \min_{\hat{X}_i} \frac{1}{2} || X_i - \hat{X}_i ||^2_{F} + (1-\lambda ) \Omega(\hat{X})
\end{equation}
Here we balance the complexity of the generated $\hat{x}$ with the mean of square distance errors by a function $\Omega$. $\lambda$  is a hyperparameter that controls the weights of recovered errors and the complexity of output.

\begin{algorithm}
\caption{Model training algorithm}\label{alg:train}
\begin{algorithmic}[1]
\State \textbf{Trainable parameters:}
\State \hskip1.0em $\Theta_{G}$ for graph components; $\Theta_{Trans}$, $\Theta_{projection}$ for transformer components
\Procedure{PredictExtremeEvents}{$X_{train}$, $X_{test}$}
    \State \textbf{build} model \GTrans and \textbf{initialize} model weights:
    \Statex \hskip11.em $\Theta = \{ \Theta_{G}$, $\Theta_{Trans}$, $\Theta_{projection} \}$
    \For{$1 \dots n_{epochs}$}
        \While{ $T$ sequences available from $X_{train}$ at time $t$ }
            \State \textbf{train} the model with 
            \Statex \hskip6.0em the input sequence $\{ X_{t-T}, X_{t-T+1}, \dots, X_{t} \}$ and
            \Statex \hskip6.0em the target sequence $\{ X_{t-T+1}, X_{t-T+2}, \dots, X_{t+1} \}$.
        \EndWhile
    \EndFor
    \State $\hat{h} \gets$ Encoder($X_{test}$) \Comment{Latent vector from Graph and Transformer encoder}
    \State $\mathbf{Err} \gets$ Decoder($\hat{h}$) - $X_{test}$
    \State $\mu \gets$ Estimate($\mathbf{Err}$) \Comment{Mahalanobis distance for Err $\in \mathbb{R}^{N\times D}$}
    \State $\epsilon \gets \lambda*\mu$
    \For{each $X$ from $X_{test}$}
        \State Err($x$) $\gets$ Decoder(Encoder($X$)) - $X$
        \State Pred($X$) $\gets \begin{cases} 0 & \text{if Err(X) $\leq \epsilon$} \\ 1 & \text{otherwise} \end{cases} $
    \EndFor
\EndProcedure
\end{algorithmic}
\end{algorithm}

We train our model \GTrans by minimizing Equation~\ref{eq:loss} using the ADAM algorithm~\cite{kingma2014adam}. The model training is self-supervised by the source input and target input. The training target window is one step ahead of the source window.
Algorithm~\ref{alg:train} gives the detailed model training and events prediction algorithm. Within $n_{epochs}$ epochs of training loop, the model is trained with $T$ length of the input sequence and generates the same length of sequences at one-step ahead.

\section{Experiments and Evaluation}
\label{sec:exp}

\subsection{Datasets}
We use three publicly available datasets for all experiments: 1) Earthquakes in Southern California, and 2) Motor vehicle collisions in New York City.  Table~\ref{tbl:data-stats} shows the basic statistics of these datasets. Figure~\ref{fig:graph-sceq} is the graph used for the selected Earthquake regions in Southern California and Figure~\ref{fig:graph-nymotor} illustrates the graph for the 45 zip-coded areas in Manhattan of New York City. The graph node connections and features pre-processing can be found in Appendix~\ref{sec:appendix}.

\begin{table}[htbp]
    \centering
    \caption{Statistics of datasets: events column means un-preprocessed events in the raw datasets, the adjacency matrix is the graph corresponding matrix, }
    \tabulinestyle{gray}
    \begin{tabu}to\columnwidth{X[1.4,l]|X[1,c]|X[1,c]|X[1,c]|X[1,c]|X[1,c]}
        \tabucline-
        Dataset & Events & Adjacency & Feature & Samples  & Extreme \\
                &        & Matrix    & size   &           & event rate \\
        \tabucline[1pt gray]-
        1) scearthquake & 609,096 & $16\times16$               & 3             &  25,515             & 9.25\% \\ \tabucline-
        2) nymotorcrash & 1,822,796 & $45\times45$               & 6             &  40,316           & 1.86\% \\ \tabucline-
    \end{tabu}
    \label{tbl:data-stats}
\end{table}

\subsubsection{Earthquakes in Southern California (scearthquake):}
This dataset contains seismic events in the Southern California area, ranging the longitude from $32^\circ$  to $36^\circ$ and latitude from $-120^\circ$ to $-116^\circ$. Seismic events are time series of data with geo locations. All locations are girded into 40 by 40 cells, each of which has .1 degree of longitude and latitude.

\subsubsection{Motor vehicle collisions in New York City (nymotorcrash):}
This dataset contains vehicle collision and crashing events. These events are time and geo location associated. 


\subsection{Experimental Setup and Implementation}

We implement the model in PyTorch~\cite{paszke2019pytorch} and train on a compute node equipped with an Intel(R) Xeon(R) CPU E5-2670 v3 @ 2.30GHz, 128GB memory and 8 NVidia K80 GPUs. For the graph convolution, we use a simplified two-layer graph convolution as~\cite{kipf2016semi}. We use PyTorch's built-in transformer layers as the backbone for building Transformer Encoder and Decoder.
We transform the original tablet data frames into time sequences of graphs. For 1) scearthquake, the graph is a $16\times16$ mesh grid by splitting the Longitude from $-120^\circ$ to $-116^\circ$ and Latitude from $+32^\circ$ to $+36^\circ$. For 2) nymotorcrash, the graph nodes are 45 zip-coded areas in the borough of Manhattan of New York City. Node connections are based on geo connections of these areas. The 3) volcanic dataset consists of 4431 natural regional segments. Hence, the graph is a complete graph by assumption.

We use Mean Squared Errors (MSE) for computing the loss in Equation~\ref{eq:loss}. We use the ADAM optimizer~\cite{kingma2014adam} with the learning rate decay when there is no improvement for the five continuous previous epochs. The statistics of all datasets, such as the adjacency matrix, feature size, number of samples, train and test splits, are shown in Table~\ref{tbl:data-stats}.


\subsection{Model Performance Evaluation}\label{sec:perf-measure}

To compare the model performance, we build three sets of baseline models: 1) MLP-AE, 2) LSTM-AE, 3) GCN-LSTM. MLP-AE is a model with a time distributed encoder, a linear projection layer, and a time distributed decoder. All the layers in this model are fully-connected dense neural layers. The architecture of this model is to simulate the same component connection as \GTrans. Similarly, LSTM-AE is a typical RNN encoder decoder model from~\cite{cho2014learning}. It adopts the same model structures except that the encoder consists of stacked LSTM layers and the decoder consists of LSTM layers in which the hidden dimensions are mirrored to the encoder. In LSTM-AE, the encoder encodes the input over time and produces a future hidden state. The decoder repeats the encoded states and produces a decoded output. Compared to LSTM-AE, GCN-LSTM replaces the encoder with a two-layer graph convolutional network and adopts the same time-wise graph convolutional decoder as \GTrans. The hidden states are propagated using one LSTM layer.

\subsubsection{Threshold}
Throughout all experiments, positive results are the extreme events that we are interested in prediction. We use a threshold value to determine whether positive or negative. We calculate the threshold in the following procedure. After models learn the latent space of datasets, we calculate the reconstruction error for each sample $Err = |X-\hat{X}| \rightarrow \mathbb{R}^{N\times D}$. We estimate the mean vector $\mathbf{\mu}$ and covariance matrix $\mathbf{\Sigma}$ from all flattened errors $\mathbb{R}^{N \cdot D}$. Then, we calculate the error distance by Mahalanobis distance~\cite{de2000mahalanobis}. Finally, the threshold value is pulled from all error distances that is proportional to the extreme rates.

\subsubsection{Performance Measures}

\begin{table}[htbp]
    \centering
    \caption{Model performance comparison. Columns of the table, such as True-Positive Rate (TPR), Accuracy (ACC), F$_1$ (F$_1$-score), and F$_2$ (F$_1$-score) are defined in Section~\ref{sec:perf-measure}.}
    \label{tbl:comp-results}
    \tabulinestyle{gray}
    \begin{tabu}to\columnwidth {X[2.5,l]|X[2.1,l]|X[0.9,c]|X[0.9,c]|X[0.72,c]|X[0.72,c]|X[1.3,c]|X[1.3,c]|X[1.3,c]|X[1.3,c]}
    \tabucline-
    \textbf{Dataset} & \textbf{Model} & \textbf{TN} & \textbf{FP} & \textbf{FN} & \textbf{TP} & \textbf{TPR} & \textbf{ACC} & \textbf{F1} & \textbf{F2}\\
    \tabucline[1pt gray]-
    \multirow{4}{*}{ scearthquake } & MLP-AE & 2019 & 2038 & 67 & 129 & \textbf{0.6582} & 0.5051 & 0.1092 & 0.2186 \\ \tabucline{2-10}
            & LSTM-AE        & 3572 & 393  & 221 & 67  & 0.2326          & \textbf{0.8556} & 0.1791          & 0.2078          \\\tabucline{2-10}
            & GCN-LSTM       & 2572 & 1427 & 139 & 115 & 0.4528          & 0.6318          & 0.1281          & 0.2248          \\ \tabucline{2-10}
            & \GTrans (ours) & 3109 & 890  & 131 & 123 & 0.4843          & 0.7599          & \textbf{0.1942} & \textbf{0.3031} \\
     \tabucline-
    \multirow{4}{*}{ nymotorcrash } & MLP-AE & 3802 & 2778 & 58 & 82  & 0.5857          & 0.5780 & 0.0547 & 0.1199 \\ \tabucline{2-10}
            & LSTM-AE        & 4708 & 1857 & 96  & 59  & 0.3806          & \textbf{0.7094} & 0.0570          & 0.1163          \\ \tabucline{2-10}
            & GCN-LSTM       & 3049 & 2347 & 470 & 854 & \textbf{0.6450} & 0.5808          & 0.3775          & 0.5025          \\ \tabucline{2-10}
            & \GTrans (ours) & 3255 & 2038 & 526 & 901 & 0.6314          & 0.6185          & \textbf{0.4127} & \textbf{0.5210} \\
    \tabucline-
    \end{tabu}
\end{table}

We compare \GTrans with the baseline models on the datasets with performance metrics. The results are listed in Table~\ref{tbl:comp-results}, where TN, FP, FN, and TP represent true negative, false positive, false negative, and true positive respectively. TPR is the true-positive rate defined as $TP/(TP+FN)$. ACC means the accuracy defined as $(TP+TN)/Total$. F$_1$ score is calcuated as $TP/(TP + 1/2(FP+FN))$. F$_2$ score focuses more on the true positive rate, calcuated as $TP/(TP+0.2FP+0.8FN)$. We train all the models on 21,262 samples and test 4,253 samples for the scsearchquake dataset, and train on 33,596 samples and test on 6,720 samples for the nymotorcarsh dataset. \GTrans achieves the highest F$_1$ and F$_2$ scores for the two datasets. Event though MLP-AE and LSTM-AE hit the highest TPR and ACC respectively for the scearthquake, F$_1$ score is low for MLP-AE and F$_2$ is the lowest for LSTM-AE. This is a similar situation for the nymotorcrash, where LSTM-AE and GCN-LSTM get the highest scores in ACC and TPR respectively. Compared to MLP-AE and LSTM-AE, graph embeddings empower both GCN-LSTM and \GTrans to achieve higher F$_2$ scores. 

\subsection{Ablation Study and t-SNE Visualization}


\begin{figure}[htpb]
    \centering
    \subfloat[T-SNE with LSTM-AE encoder \label{fig:tsne-lstmae}]{
        \includegraphics[width=0.475\textwidth]{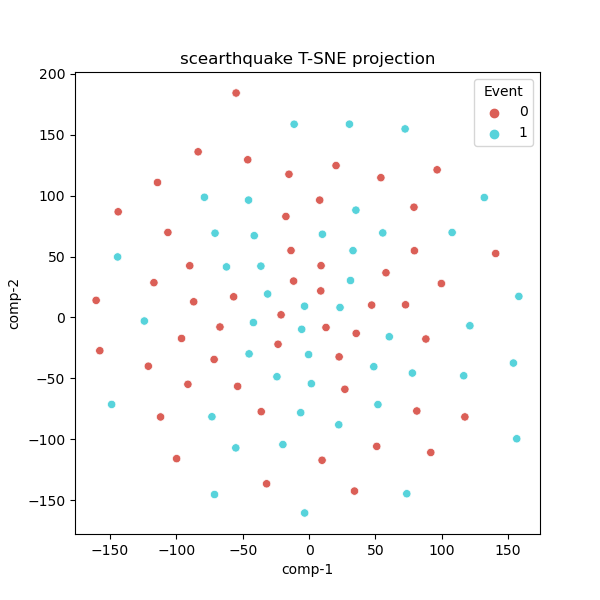}
    }\hfill
    \subfloat[T-SNE with \GTrans encoder \label{fig:tsne-gtrans}]{
        \includegraphics[width=0.475\textwidth]{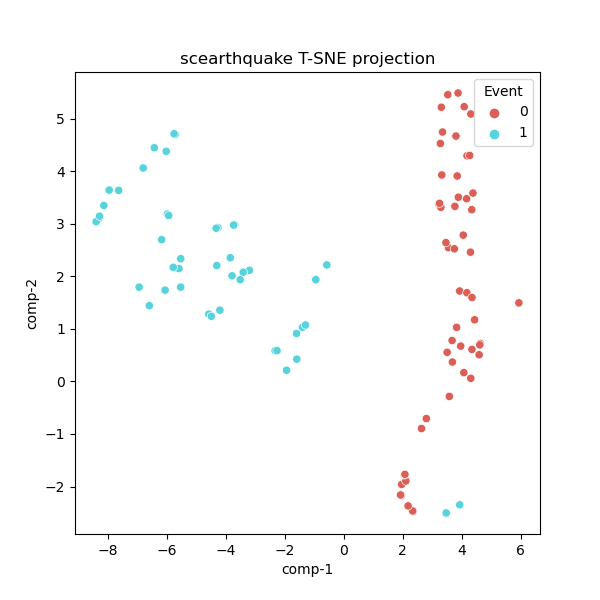}
    }
    \caption{T-SNE visualization for the latent space from LSTM-AE and \GTrans using 100 events from the training datasets.}
    \label{fig:tsne}
\end{figure}

GCN-LSTM and \GTrans share the same size of parameters for the graph encoder and decoder. We swap the transformer components with two LSTM layers as shown as the baseline GCN-LSTM. For the results in Table~\ref{tbl:comp-results}, \GTrans has a 3\% better TPR and 12\% more ACC for the scearthquake. For the nymotorcrash, the TPR is almost the same for both and \GTrans has 3\% more ACC score. In addition, with Transformer components, \GTrans is able to gain more benefits from training with longer sequences. Compared to the testing metrics, training \GTrans on sequences with 100 window size for the nymotorcrash achieves about 9\% higher scores, whereas the LSTM-based LSTM-AE and GCN-LSTM can achieve 4 and 5\% improvement using longer sequences. 


Figure~\ref{fig:tsne} visualizes the latent space using t-Distributed Stochastic Neighbor Embedding (t-SNE)~\cite{van2008visualizing} from LSTM-AE and \GTrans. All the two sub-figures have 100 events from the training dataset. T-SNE algorithm reduces the high dimensional latent space into two major components for visualization. T-SNE is trained with a random initialized centroids and set the learning rate following from~\cite{kobak2019art}. Figure~\ref{fig:tsne-lstmae} is the t-SNE plot for the latent space from LSTM-AE and Figure~\ref{fig:tsne-gtrans} displays the plot from \GTrans. It is obvious that \GTrans has achieved a better separation for the two categorical data.


\section{Conclusion}
\label{sec:con}

We introduce \GTrans, a transformer-based AutoEncoder with Graph embeddings. It utilizes the symmetric graph encoder and decoder for capturing spatial structures between nodes and predicts temporal correlation by capturing long term context with Transformer encoder and decoder. We formulate the spatiotemporal prediction problem settings and illustrate the neural network components. Our experiments demonstrate the effectiveness of the model on predicting extreme rare events in a time series fashion. This model setting can be further extended to the online machine learning where the sequential order must be maintained.

%
%
%
\bibliographystyle{splncs04}
\bibliography{refs}

\begin{thebibliography}{10}
\providecommand{\url}[1]{\texttt{#1}}
\providecommand{\urlprefix}{URL }
\providecommand{\doi}[1]{https://doi.org/#1}

\bibitem{bertasius2021space}
Bertasius, G., Wang, H., Torresani, L.: Is space-time attention all you need
  for video understanding? arXiv preprint arXiv:2102.05095  (2021)

\bibitem{bordes2013translating}
Bordes, A., Usunier, N., Garcia-Duran, A., Weston, J., Yakhnenko, O.:
  Translating embeddings for modeling multi-relational data. Advances in neural
  information processing systems  \textbf{26} (2013)

\bibitem{bronstein2017geometric}
Bronstein, M.M., Bruna, J., LeCun, Y., Szlam, A., Vandergheynst, P.: Geometric
  deep learning: going beyond euclidean data. IEEE Signal Processing Magazine
  \textbf{34}(4),  18--42 (2017)

\bibitem{cho2014learning}
Cho, K., Van~Merri{\"e}nboer, B., Gulcehre, C., Bahdanau, D., Bougares, F.,
  Schwenk, H., Bengio, Y.: Learning phrase representations using rnn
  encoder-decoder for statistical machine translation. arXiv preprint
  arXiv:1406.1078  (2014)

\bibitem{de2000mahalanobis}
De~Maesschalck, R., Jouan-Rimbaud, D., Massart, D.L.: The mahalanobis distance.
  Chemometrics and intelligent laboratory systems  \textbf{50}(1),  1--18
  (2000)

\bibitem{dosovitskiy2020image}
Dosovitskiy, A., Beyer, L., Kolesnikov, A., Weissenborn, D., Zhai, X.,
  Unterthiner, T., Dehghani, M., Minderer, M., Heigold, G., Gelly, S., et~al.:
  An image is worth 16x16 words: Transformers for image recognition at scale.
  In: International Conference on Learning Representations (2021)

\bibitem{geng2019spatiotemporal}
Geng, X., Li, Y., Wang, L., Zhang, L., Yang, Q., Ye, J., Liu, Y.:
  Spatiotemporal multi-graph convolution network for ride-hailing demand
  forecasting. In: Proceedings of the AAAI conference on artificial
  intelligence. vol.~33, pp. 3656--3663 (2019)

\bibitem{kingma2014adam}
Kingma, D.P., Ba, J.: Adam: A method for stochastic optimization. arXiv
  preprint arXiv:1412.6980  (2014)

\bibitem{kipf2016semi}
Kipf, T.N., Welling, M.: Semi-supervised classification with graph
  convolutional networks. arXiv preprint arXiv:1609.02907  (2016)

\bibitem{kobak2019art}
Kobak, D., Berens, P.: The art of using t-sne for single-cell transcriptomics.
  Nature communications  \textbf{10}(1),  1--14 (2019)

\bibitem{van2008visualizing}
Van~der Maaten, L., Hinton, G.: Visualizing data using t-sne. Journal of
  machine learning research  \textbf{9}(11) (2008)

\bibitem{park2019symmetric}
Park, J., Lee, M., Chang, H.J., Lee, K., Choi, J.Y.: Symmetric graph
  convolutional autoencoder for unsupervised graph representation learning. In:
  Proceedings of the IEEE/CVF International Conference on Computer Vision. pp.
  6519--6528 (2019)

\bibitem{paszke2019pytorch}
Paszke, A., Gross, S., Massa, F., Lerer, A., Bradbury, J., Chanan, G., Killeen,
  T., Lin, Z., Gimelshein, N., Antiga, L., et~al.: Pytorch: An imperative
  style, high-performance deep learning library. Advances in neural information
  processing systems  \textbf{32},  8026--8037 (2019)

\bibitem{salinas2020deepar}
Salinas, D., Flunkert, V., Gasthaus, J., Januschowski, T.: Deepar:
  Probabilistic forecasting with autoregressive recurrent networks.
  International Journal of Forecasting  \textbf{36}(3),  1181--1191 (2020)

\bibitem{taubin1995signal}
Taubin, G.: A signal processing approach to fair surface design. In:
  Proceedings of the 22nd annual conference on Computer graphics and
  interactive techniques. pp. 351--358 (1995)

\bibitem{vaswani2017attention}
Vaswani, A., Shazeer, N., Parmar, N., Uszkoreit, J., Jones, L., Gomez, A.N.,
  Kaiser, {\L}., Polosukhin, I.: Attention is all you need. In: Advances in
  neural information processing systems. pp. 5998--6008 (2017)

\bibitem{wang2018non}
Wang, X., Girshick, R., Gupta, A., He, K.: Non-local neural networks. In:
  Proceedings of the IEEE conference on computer vision and pattern
  recognition. pp. 7794--7803 (2018)

\bibitem{wu2020deep}
Wu, N., Green, B., Ben, X., O'Banion, S.: Deep transformer models for time
  series forecasting: The influenza prevalence case. arXiv preprint
  arXiv:2001.08317  (2020)

\bibitem{yan2018spatial}
Yan, S., Xiong, Y., Lin, D.: Spatial temporal graph convolutional networks for
  skeleton-based action recognition. In: Thirty-second AAAI conference on
  artificial intelligence (2018)

\bibitem{yoon2019time}
Yoon, J., Jarrett, D.: Time-series generative adversarial networks. In: 33rd
  Conference on Neural Information Processing Systems ({NeurIPS} 2019). p.~11
  (2019)

\bibitem{zerveas2021transformer}
Zerveas, G., Jayaraman, S., Patel, D., Bhamidipaty, A., Eickhoff, C.: A
  transformer-based framework for multivariate time series representation
  learning. In: Proceedings of the 27th ACM SIGKDD Conference on Knowledge
  Discovery \& Data Mining. pp. 2114--2124 (2021)

\bibitem{zhu2020adversarial}
Zhu, S., Yuchi, H.S., Xie, Y.: Adversarial anomaly detection for marked
  spatio-temporal streaming data. In: ICASSP 2020-2020 IEEE International
  Conference on Acoustics, Speech and Signal Processing (ICASSP). pp.
  8921--8925. IEEE (2020)

\end{thebibliography}
\clearpage
\pagebreak
\appendix
\section{Appendix}\label{sec:appendix}

\subsection{Data pre-processing}

\begin{figure}[htpb]
    \centering
    \subfloat[Graph of Earthquake regions in Southern California\label{fig:graph-sceq}]{
        \includegraphics[width=0.475\textwidth]{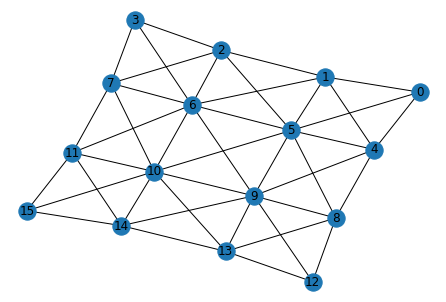}
    }\hfill
    \subfloat[Graph of the Borough of Manhattan, New York City\label{fig:graph-nymotor}]{
        \includegraphics[width=0.475\textwidth]{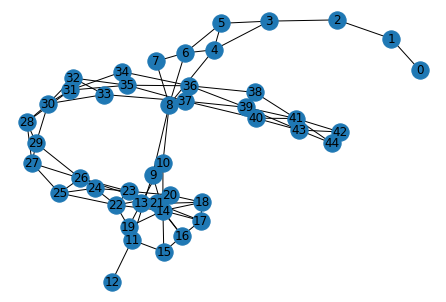}
    }
    \caption{Graph of the datasets used in the graph components from \GTrans and GCN-LSTM: a) scearthquake, b) nymotorcrash. For 1) scearthquake, there are 16 nodes; for 2) nymotorcrash, there are 45 nodes.}
    \label{fig:graphs}
\end{figure}

The two datasets are public available. For 1) scearthquake, the dataset can be downloaded from the USGS website\footnote{\url{https://www.usgs.gov/}}. And 2) nymotorcrash dataset can be found from the NYC open data project\footnote{\url{https://opendata.cityofnewyork.us/}}.
For 1) scearthquake, there are 16 regions according to geo longitudes and latitudes, where each node represents one region and the connections are based on the natural connection of regions; for 2) nymotorcrash, there are 45 zip-coded areas, each of which is connected the road traffic connected areas.
The node connections are equally weighted at initialization. The features used in the scearthquake are the magnitude, the depth, and the significance. The features used in nymotorcrash are the number of people injured or killed. All features are 0-1 normalized before training.


\setcounter{tocdepth}{1}
\listoftodos

\end{document}